\documentclass[11pt]{article}
\usepackage{fullpage}
\makeatletter
\long\def\@makecaption#1#2{
  \vskip 0.8ex
  \setbox\@tempboxa\hbox{\small {\bf #1:} #2}
  \parindent 1.5em 
  \dimen0=\hsize
  \advance\dimen0 by -3em
  \ifdim \wd\@tempboxa >\dimen0
  \hbox to \hsize{
    \parindent 0em
    \hfil 
    \parbox{\dimen0}{\def\baselinestretch{0.96}\small
      {\bf #1.} #2
     
    } 
    \hfil}
  \else \hbox to \hsize{\hfil \box\@tempboxa \hfil}
  \fi
}
\makeatother

\usepackage{times,enumitem,hyperref,booktabs,multicol,blindtext,url,parskip}

\usepackage{pifont}

\newcommand{\cmark}{\ding{51}}%
\newcommand{\xmark}{\ding{55}}%

\usepackage{amsfonts}
\usepackage{amsmath}
\usepackage{amsthm}
\usepackage{amssymb}
\usepackage{caption}
\usepackage{microtype}
\usepackage{graphicx}
\usepackage{subcaption}
\usepackage{natbib}
\usepackage{fancyhdr}
\usepackage{color}
\usepackage{algorithm}
\usepackage{forloop}
\usepackage{amsthm}
\usepackage[english]{babel}
\usepackage[T1]{fontenc}
\usepackage{amsmath}
\usepackage{amssymb}
\usepackage{amsfonts}
\usepackage{multirow}
\usepackage{mathtools}
\usepackage{breqn}
\usepackage{bbm}
\usepackage{dsfont}
\usepackage{graphicx}
\usepackage{natbib}
\usepackage{cases}
\usepackage[colorinlistoftodos]{todonotes}

\renewcommand{\hat}{\widehat}

\newtheorem*{remark*}{Remark}

\newtheorem*{observation*}{Observation}

\numberwithin{equation}{section}

\newcommand{\Gnorm}[1]{{\left\vert\kern-0.25ex\left\vert\kern-0.25ex\left\vert #1 
		\right\vert\kern-0.25ex\right\vert\kern-0.25ex\right\vert}}
\newcommand{\gnorm}[1]{{\vert\kern-0.25ex\vert\kern-0.25ex\vert #1 
		\vert\kern-0.25ex\vert\kern-0.25ex\vert}}

\usepackage{multirow}
\usepackage{enumerate}
\usepackage{algpseudocode}
\usepackage{rotating}
\usepackage{makecell}
\usepackage{bm}

\def\ie{\textit{i.e.}}
\def\eg{\textit{e.g.}}

\usepackage[capitalize]{cleveref}
\crefname{section}{Sec.}{Secs.}
\Crefname{section}{Section}{Sections}
\Crefname{table}{Table}{Tables}
\crefname{table}{Tab.}{Tabs.}

\def\shownotes{0}  %set 1 to show author notes
\ifnum\shownotes=1
\newcommand{\authnote}[2]{[#1: #2]}
\else
\newcommand{\authnote}[2]{}
\fi

\definecolor{mypink}{RGB}{219, 48, 122}

\usepackage{xcolor}

\hypersetup{
  colorlinks,
  citecolor=violet,
  linkcolor=red,
  urlcolor=blue}

\begin{document}

% Control whitespace around equations
\abovedisplayskip=8pt plus0pt minus3pt
\belowdisplayskip=8pt plus0pt minus3pt

\begin{center}
  \vspace*{-1cm}
  {\LARGE Federated Generalized Category Discovery  \\
  \vspace{0.12cm}
  ~} \\
  \vspace{.4cm}
  {\Large Nan Pu$^{\textcolor{mypink}{1}}$ ~~~~ Zhun Zhong$^{\textcolor{mypink}{1}}$ 
  ~~~~ Xinyuan Ji$^{\textcolor{mypink}{2}}$
  ~~~~ Nicu Sebe$^{\textcolor{mypink}{1}}$ 
  } \\
  \vspace{.4cm}
  \textcolor{mypink}{$^1$} University of Trento 
  \textcolor{mypink}{$^2$} Xi'an Jiaotong University
  
\end{center}

\begin{abstract}
Generalized category discovery (GCD) aims at grouping unlabeled samples from known and unknown classes, given labeled data of known classes. To meet the recent decentralization trend in the community, we introduce a practical yet challenging task, namely Federated GCD (Fed-GCD), where the training data are distributively stored in local clients and cannot be shared among clients. The goal of Fed-GCD is to train a generic GCD model by client collaboration under the privacy-protected constraint. The Fed-GCD leads to two challenges: 1) representation degradation caused by training each client model with fewer data than centralized GCD learning, and 2) highly heterogeneous label spaces across different clients. To this end, we propose a novel Associated Gaussian Contrastive Learning (AGCL) framework based on learnable GMMs, which consists of a Client Semantics Association (CSA) and a global-local GMM Contrastive Learning (GCL). On the server, CSA aggregates the heterogeneous categories of local-client GMMs to generate a global GMM containing more comprehensive category knowledge. On each client, GCL builds class-level contrastive learning with both local and global GMMs. The local GCL learns robust representation with limited local data. The global GCL encourages the model to produce more discriminative representation with the comprehensive category relationships that may not exist in local data. We build a benchmark based on six visual datasets to facilitate the study of Fed-GCD. Extensive experiments show that our AGCL outperforms the FedAvg-based baseline on all datasets.
\end{abstract}

\section{Introduction}

\begin{figure}[ht]
  \begin{center}

  \includegraphics[width=0.8\textwidth]{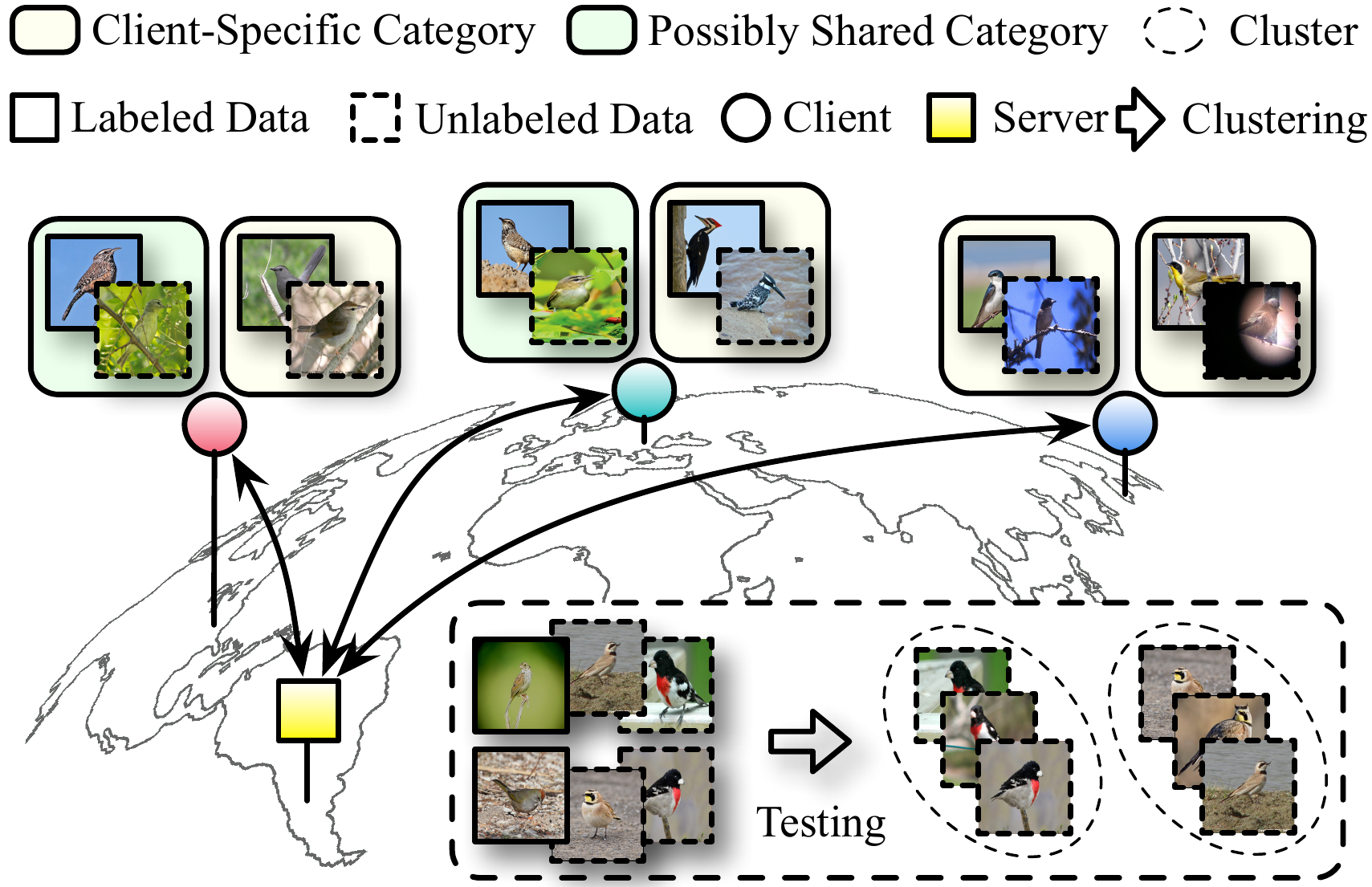}
\caption{Illustration of the proposed Fed-GCD with the case of global bird species discovery. In Fed-GCD, the data are distributively collected from the different local stations (clients) over the world, which are partially annotated. Each client includes client-specific categories and may share some common categories with the other clients.
Moreover, the raw data stored in local clients are not allowed to share with the central server or other clients, due to data privacy. 
The goal of Fed-GCD is to collaboratively train a generic GCD model under the federated privacy constraint, and then utilize it to discover novel categories in the unlabeled data on clients or the server during testing.
\label{fig:conceptual}}
  \end{center}
\end{figure}
Generalized category discovery (GCD) seeks to categorize unlabeled samples from known and unknown classes by leveraging labeled data of known classes. As a more practical extension of novel category discovery (NCD)~\cite{han2019learning, fini2021unified,joseph2022novel,zhong2021neighborhood,zhong2021openmix,han2021autonovel,zhang2022grow,roy2022class,yu2022self,DBLP:conf/iclr/ChiLYLL00ZS22}, GCD has attracted increasing attention.
While existing GCD methods~\cite{vaze2022generalized,DBLP:conf/bmvc/FeiZYZ22,yang2022divide,sun2022opencon} have achieved promising performance, they always require centralized training, where the training data need to be accessed at once. However, this strategy violates many practical application scenarios: the GCD data are distributively collected by different local clients and the data in each client cannot be shared with others due to the privacy concerns. For instance, as shown in~\cref{fig:conceptual}, a global species research center plans to discover the new species of global birds through the collaboration of local stations located around the world.
Each local station is responsible for capturing and partially annotating bird images. Due to the difference in local policies and laws, it is hard to make an agreement to share the local data between stations. Thus, a decentralized system is required to handle this pragmatic GCD scenario.
\par

To meet this requirement, we propose a practical yet challenging task, namely Federated GCD (Fed-GCD), in which the GCD data are individually collected and partially annotated by local clients as well as cannot be shared with other clients. The objective of Fed-GCD is to train a generic GCD model via the collaboration across local clients without sharing local samples, which can recognize both known and unknown categories in the unlabeled data. Compared with the conventional federated learning (FL) setups~\cite{DBLP:conf/iclr/LiJZKD21,DBLP:conf/iclr/Diao0T21,DBLP:conf/mlsys/LiSZSTS20,DBLP:conf/iclr/AcarZNMWS21,DBLP:conf/nips/KhodakBT19,DBLP:journals/corr/abs-1910-03581}, in Fed-GCD, local data are partially-labeled and unlabeled data may belong to unknown categories that disappear in labeled data. In addition, clients may share some common categories since some species of birds could live in different continents as shown in \cref{fig:conceptual}, and the different clients may have distinct client-specific categories. Attributed to such a complicated yet real situation, Fed-GCD suffers from 1) additional difficulties caused by open-set learning on limited local data, and 2) more severe data heterogeneity problems due to the inconsistent label space between clients. 
\par
To tackle the challenges in Fed-GCD, we propose a novel Associated Gaussian Contrastive Learning (AGCL) framework, which unifies the discriminative representation learning on the limited local data and the heterogeneous category aggregation on the central server, benefiting from learnable GMMs. Specifically, we propose to represent the potential classes by a learnable Gaussian mixture model (GMM), which brings two advantages. First, the learnable mechanism enables us to perform class-aware contrastive learning with dynamic Mahalanobis distance, which can reduce the side effects of inaccurate clustering. Second, modeling the classes as GMMs is favorable for generating informative feature-level samples of each category on server, without assessing the raw data. 

To this end, we propose a client semantics association (CSA) on the central server and a global-local GMM Contrastive Learning (GCL) on local clients. CSA builds a new feature set by sampling from each category of the uploaded local GMMs generated by clustering local data.
Then, CSA aggregates the category knowledge by clustering on the feature set, which yields a global GMM. This process not only implicitly aligns the shared classes across local clients, but also aggregates client-specific category information. As a result, the global GMM can enrich both the intra- and inter-class relationships for local training. GCL targets at performing robust contrastive representation learning by jointly using global and local GMMs. On the one hand, the GMM-based contrastive learning is insensitive to wrongly pseudo-labeled samples, which can help the model to learn robust representation. On the other hand, the association of global-local GMMs enforces the model to learn more generalized representation in a complementary way.  
\par
We summarize the contributions of this work as follows:
\begin{itemize}
    \item \textbf{Task contribution.} We explore a new yet practical GCD task, namely Fed-GCD, which investigates GCD problems under a federated learning scenario.
    
    \item \textbf{Technical contribution.} We propose a new AGCL framework for Fed-GCD. AGCL fully takes the advantage of GMMs to learn generalized representation in a robust and comprehensive manner.

    \item \textbf{Empirical contribution.} We build a Fed-GCD benchmark with different degrees of data heterogeneity based on six datasets to simulate possible conditions in real-world GCD applications. Experiments demonstrate that the proposed AGCL can improve performance across all settings.

\end{itemize}

\section{Related Work}
\par\noindent
\textbf{Generalized Category Discovery} (GCD) aims to categorize all images in an unlabelled set by using the knowledge learned from a set of labeled categories. Unlike earlier related tasks such as Novel Category Discovery~\cite{han2019learning,fini2021unified,zhao2021novel, 2211.11727} (NCD) and generalized transfer learning~\cite{hsu2018learning,hsu2018multi}, GCD assumes that the unlabeled data comes from both known and unknown categories. Therefore, GCD is a practical and challenging task that has attracted increasing attention~\cite{vaze2022generalized,zhang2022grow,DBLP:conf/bmvc/FeiZYZ22,joseph2022novel,roy2022class,joseph2022novel,yang2022divide,yu2022self,DBLP:conf/iclr/ChiLYLL00ZS22,sun2022opencon,zhong2021neighborhood,han2021autonovel,2305.06144}. For example, GCD~\cite{vaze2022generalized} has indicated that the combination of self-supervised and supervised representation learning is helpful for improving clustering discovery. XCon~\cite{DBLP:conf/bmvc/FeiZYZ22} has proposed learning with multiple experts for fine-grained category discovery. OpenCon~\cite{sun2022opencon} has demonstrated the significant superiority of jointly considering prototypical contrastive learning and pseudo-label assignment. \textit{Although these methods show promising performance under relatively practical assumptions, they neglect the increasingly important issue of data privacy. To investigate this overlooked issue and address additional technical bottleneck, we design a Fed-GCD task and introduce a new AGCL framework accordingly.}

\par\noindent
\textbf{Federated Learning} (FL) as a promising solution for privacy-preserving decentralized training was first introduced in~\cite{mcmahan2017communication}. In the typical FL algorithm, FedAvg~\cite{mcmahan2017communication}, the goal is to learn a global model by averaging weight parameters across local models trained on private client datasets. Most existing FL works~\cite{DBLP:conf/iclr/LiJZKD21,DBLP:conf/iclr/Diao0T21, DBLP:conf/mlsys/LiSZSTS20,DBLP:conf/iclr/AcarZNMWS21,DBLP:conf/nips/KhodakBT19,DBLP:journals/corr/abs-1910-03581} focus on supervised learning settings, where the local private data are fully labeled. However, the assumption that all of the data examples are fully annotated is not realistic for real-world applications like GCD. Thus, one early work~\cite{2010.08982} has attempted to introduce self-supervised learning into the FL framework. Later, since there is often partially-labeled data in real-world scenarios, some semi-supervised FL approaches~\cite{DBLP:journals/corr/abs-2108-09412,2205.13921} are proposed to exploit the partial supervision and learn better representations with few annotation costs. As summarized in~\cref{tab:fl_setup}, these works assume local clients share a common label space that is infeasible for GCD tasks. \textit{In contrast, our Fed-GCD is challenged by more severe issues of data heterogeneity, because the label space on clients may be non-overlapping or clients share only a few classes with each other.} 

\par\noindent
\textbf{Contrastive learning} (CL)~\cite{chen2020simple} has been demonstrated to be highly effective for representation learning in a self-supervised setting. Inspired by the powerful CL approaches~\cite{jaiswal2020survey,he2020momentum,chuang2020debiased,khosla2020supervised}, GCD~\cite{vaze2022generalized} has introduced a combination of the self-supervised and the semi-supervised learning to enhance GCD representation. Moreover, prototypical contrastive learning~\cite{DBLP:conf/iclr/0001ZXH21} (PCL) further considers class-level supervision by contrasting instance features with a set of prototypes. However, PCL needs an instance-level memory buffer to produce the prototype set, which is computationally and memory-intensive. \textit{In contrast to the PCL that focus on the learning of prototypes, our GCL considers additional class-aware variances to comprehensively model data distributions without instance buffer, by incorporating the classical GMM model and contrastive learning in a unified framework. This allows models to be insensitive to outliers, especially for unreliable clusters.}

\begin{table}[!t]
\centering
\caption{Comparison between different federated learning (FL) setups. ``FS'', ``SS'' and ``SE'' denote fully-supervised, self-supervised and semi-supervised, respectively. \label{tab:fl_setup} 
}
\resizebox{0.8\textwidth}{!}{
\begin{tabular}{l|c|l}
\hline
\multicolumn{1}{c|}{FL Setup} & Out of Category Distribution & Annotation on Client \\ \hline
FS & \xmark& Fully Labeled      \\
SS \cite{2010.08982} & \xmark & Unlabeled          \\
SE \cite{DBLP:journals/corr/abs-2108-09412,2205.13921} & \xmark & Partially Labeled       \\
Fed-GCD & \cmark & Partially Labeled       \\ \hline
\end{tabular}
}
\end{table}

\section{Federated Generalized Category Discovery}
\subsection{Problem Definition and Formulation}\label{sec:Fed-GCD_problem}

Given the practical requirements of generalized category discovery (GCD) applications (\eg, species distribution and data privacy), it is necessary to build a generic GCD model via collaborative decentralized training across clients without sharing their local data. To meet these requirements, we propose a federated generalized category discovery (Fed-GCD) task. In Fed-GCD task, the local training data collected by each client are partially labeled, where the labeled data belong to known categories, and the unlabeled data may come from known or unknown novel categories. Additionally, each client learns on its distinct label set, which contains client-specific categories and may include some shared common categories. Compared to the semi-supervised federated learning~\cite{DBLP:conf/iclr/JeongYYH21} (semi-FL) setting that assumes both labeled and unlabeled data belong to known categories and share a common label space, Fed-GCD is more challenging due to highly-heterogeneous data issues attributed to inconsistent label spaces between clients and additional difficulties caused by open-set learning on local data. In light of this, Fed-GCD aims to 1) improve the local GCD model's representation learning ability on limited local data in open-set learning scenarios, and 2) associate the heterogeneous local label spaces to provide comprehensive category knowledge for local training. To the best of our knowledge, we are the first to explore the FL setup in GCD.

\par
Formally, in the Fed-GCD task, there are $N^{L}$ local client models $\{\Theta^{L}_{n}\}^{N^{L}}_{n=1}$ and one central server with the GCD model $\Theta^{G}$. In the beginning, the global model $\Theta^{G}_0$ is initialized with the weights pre-trained on a publicly available large dataset (\eg, ImageNet~\cite{deng2009imagenet}) and distributed to each client. Given the local dataset on $n$-th client $\mathcal{D}^{L}_{n}=\{(\boldsymbol{x}_{i},y_{i})\}_{i=1}^{N^{L}_{n}} \in \mathcal{X}^{L}_{n} \times \mathcal{Y}^{L}_{n}$ with the corresponding image set $\mathcal{X}^{L}_{n}$ and label set $\mathcal{L}^{L}_{n}$, the $n$-th client is required to train its local model $\Theta^{L}_{n}$ based on the distributed global model $\Theta^{G}_0$ by leveraging its local dataset $\mathcal{D}^{L}_{n}$. In our Fed-GCD setup, we assume that for $i$-th and $j$-th client, $i\neq j$,  $\mathcal{L}^{L}_{i}$ and $ \mathcal{L}^{L}_{j}$ might be partially overlapping or completely non-overlapping, but their label space cannot be same (\ie, $\mathcal{L}^{L}_{i}\bigcup\mathcal{L}^{L}_{j} \neq \mathcal{L}^{L}_{i} or \mathcal{L}^{L}_{j}$). To simulate such data distribution that often exists in real-world GCD applications, we adopt the parametric Dirichlet distribution \cite{hsu2020federated} to control the degree of data heterogeneity.

\begin{figure*}[t]
\centering
  \includegraphics[width=0.95\textwidth]{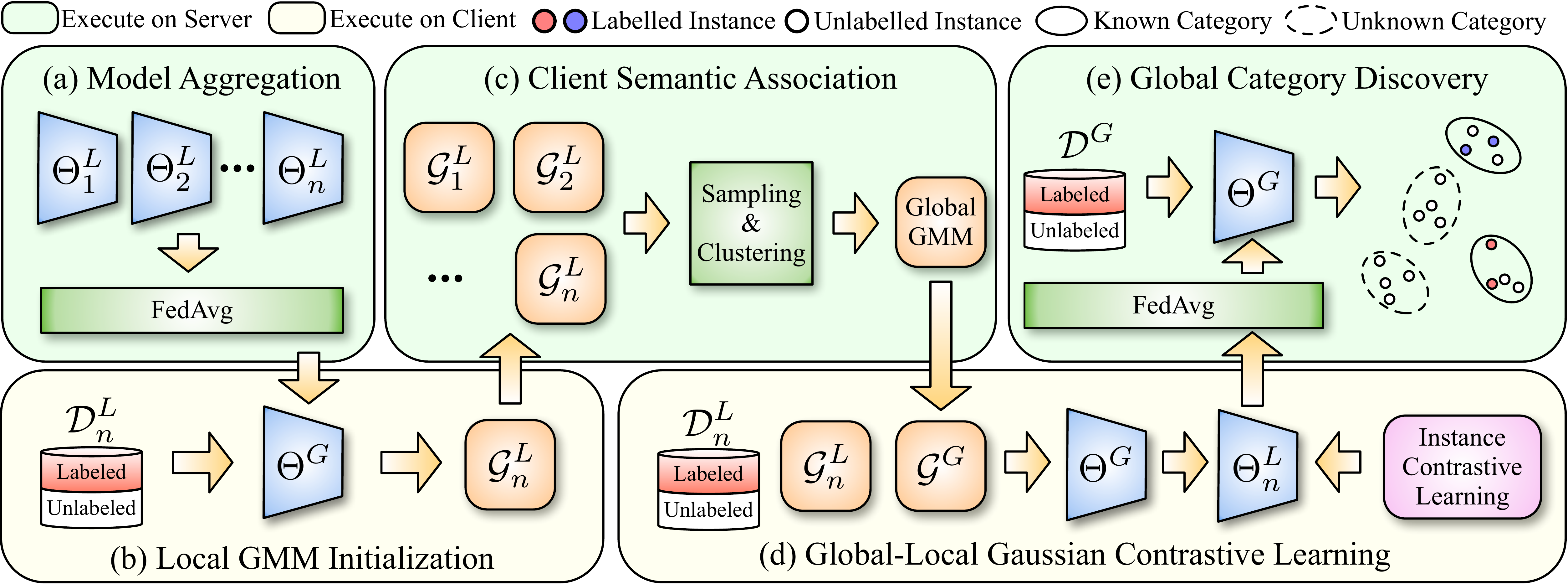}
\caption{Diagram of the proposed federated Gaussian contrastive learning (FGCL) framework. We first apply FedAvg~\cite{mcmahan2017communication} to aggregate the uploaded local models, resulting in a global model that will be distributed to all clients. Then, after leveraging the distributed models to extract image features, local clients are required to cluster these features and initialize local GMMs. Next, the local GMMs are uploaded to the central server, and aggregated by the proposed CSA, to generate a global GMM before local training. Later, the server distributes the global GMM to each client. Based on the global-local GMMs, client models are collaboratively optimized by the proposed GCL. Finally, a generic model is trained for global category discovery.
\label{fig:framework}}

\end{figure*}

\subsection{Baseline}\label{sec:baseline}
We employ the commonly-used FedAvg~\cite{mcmahan2017communication} algorithm, as our basic framework. Due to the inconsistent label spaces between local clients, we follow the previous FL work~\cite{li2021meta} that only sends the feature extractor to the server. Given a feature extractor $f$ parameterized by $\Theta$, the extracted representation is defined as $\boldsymbol{v} = f(\boldsymbol{x})$. As illustrated in~\cref{fig:framework}~(a), (d) and (e), the steps of the baseline for collaborative training by server and clients are as follows. 
\par\noindent
\textbf{Step I.} In the $t$-th communication round, the server first aggregates the client models $\Theta^{L}_{i}$ uploaded from the last communication round, by taking a weighted average of them: 
\begin{equation}
    \Theta^{G}_{t+1}= \sum^{N^{L}}_{n = 1} \frac{N^{L}_{n}}{N} \cdot \Theta^{L}_{n},~~~~ N=\sum_{i=1}^{N^{L}}N^{L}_{i}.
\end{equation}
\par\noindent
Then, the averaged model is distributed to each client.

\par\noindent
\textbf{Step II.} Based on the received global model, the $i$-th client trains its model by using local data $\mathcal{D}^{L}_{i}$ with the instance contrastive learning loss $\mathcal{L}^{I}$ proposed in~\cite{vaze2022generalized} (in \cref{fig:framework} (d)). Specifically, we define that $x_{i}$ and $\hat{x}_{i}$ are two views of random augmentations for the same image in a mini-batch $\mathcal{B} = \mathcal{B}^{L}\cup\mathcal{B}^{U}$, consisting of the labeled subset $\mathcal{B}^{L}$ and unlabeled subset $\mathcal{B}^{U}$. The extracted representation $\mathbf{v}_{i}$ is further projected by a MLP projection head $h$ to high-dimensional embedding space for instance-level contrastive learning. The loss function is formulated as: 

\begin{equation}
\begin{aligned}
&\mathcal{L}^{n}_{ins}=(\lambda-1)\sum_{i \in \mathcal{B}}\log \frac{\mathcal{S}_{ins}(\boldsymbol{v}_{i}, \hat{\boldsymbol{v}}_{i}, \tau^{S})}{\sum_{j\in \mathcal{B}, j \neq i} \mathcal{S}_{ins}(\boldsymbol{v}_{i}, \boldsymbol{v}_{j}, \tau^{S})}\\
&\sum_{i \in \mathcal{B}^{L}}\frac{-\lambda}{|\mathcal{P}(i)|} \sum_{p \in \mathcal{P}(i)} \log \frac{\mathcal{S}_{ins}(\boldsymbol{v}_{i}, \boldsymbol{v}_{p}, \tau^{L})}{\sum_{j\in \mathcal{N}(i)} \mathcal{S}_{ins}(\boldsymbol{v}_{i}, \boldsymbol{v}_{j}, \tau^{L})},\\
\end{aligned}
\end{equation}
\begin{equation}\label{eq:ins_similarity}
\mathcal{S}_{ins}(\boldsymbol{v}, \hat{\boldsymbol{v}}, \tau) = \exp \left(h\left(\boldsymbol{v}\right) \cdot h\left(\hat{\boldsymbol{v}}\right) / \tau\right),
\end{equation}
where $\mathcal{P}(i)$ and $\mathcal{N}(i)$ are the positive and the negative index set for the anchor image $i\in \mathcal{B}^{L}$, respectively. $\lambda$ is a trade-off factor to balance the contributions of self-supervised and supervised learning.

\par
\par\noindent
\textbf{Step III.} The updated global model will be transmitted to each client. \textbf{Step I} and \textbf{II} are repeated until convergence. Ultimately, we use the final global model to discover new categories (in \cref{fig:framework} (e)).

\subsection{Limitations and Motivations}
Although the baseline approach works on our Fed-GCD benchmark, it shows unsatisfactory performance compared with centralized training, especially on fine-grained GCD datasets (see \cref{tab:fine_grained}). We argue that the main reasons are attributed to two aspects: 1) the GCD~\cite{vaze2022generalized} applied in local client training mainly focuses on instance-level contrastive learning while it neglects class-level contrastive learning, especially on unlabeled data. Since class-level or prototypical supervision plays an important role in open-set learning~\cite{sun2022opencon}, the Fed-GCD fails to collaboratively train a robust global GCD model without discriminative local models; 2) sharing only the backbone network is inefficient to leverage the comprehensive category relationship that may not be observed in local clients. Moreover, although the label space of each client in Fed-GCD might potentially share some common semantic information (\eg, a specie of bird distributed on different continents), the server has no explicit knowledge to align or leverage such class-level relationships under privacy protection constraints.
\par
To overcome these limitations, we consider representing the class-level knowledge by a learnable Gaussian mixture model (GMM), which is initialized by a parameter-free clustering approach. Each component of the GMM models a potential class/cluster with class-specific mean and variance, which naturally results in a concentration-based distance metric for robust contrastive learning. This idea enables models to 1) mitigate the negative effects caused by inaccurate clustering and enforce class-level supervision into local training, and 2) generate informative feature-level samples of each category for knowledge aggregation on the server without leaking original data.

\section{Federated Gaussian Contrastive Learning}
Based on the above analyses, we propose a novel Associated Gaussian contrastive learning (AGCL) framework to accomplish efficient Fed-GCD.
AGCL consists of a global-local GMM Contrastive Learning (GCL) on local clients and a client semantics association (CSA) on the central server. The former enforces a class-level contrastive learning in local training by jointly using a global GMM and a local one, where the local GMM is created by clustering on local data and the global GMM is distributed from the central server. The latter serves to aggregate heterogeneous category knowledge contained in the local GMMs following a client-agnostic manner, and generates the global GMM to provide comprehensive category relationship for local training. The goal of AGCL is to improve representation learning by enforcing class-aware GCL and associating related semantic knowledge scattered across clients.

\subsection{Gaussian Contrastive Learning}
As empirically demonstrated in \cite{DBLP:conf/bmvc/FeiZYZ22,2208.00979,DBLP:conf/iclr/0001ZXH21}, class-level or prototypical contrastive learning is efficient for learning a clustering-friendly representation. Recently, open-set contrastive learning~\cite{sun2022opencon} further indicates that such representation learning can significantly improve the GCD model's abilities to discover both known and unknown categories. However, these methods represent a class by using only the center or the mean of the class, which is insufficient and vulnerable to wrong pseudo-labeling caused by inaccurate clustering. To address this issue, we propose to employ a classical Gaussian mixture model (GMM) to model potential cluster distributions, and then perform class-level contrastive learning across the components of the GMM. 

\par\noindent
\textbf{Revisiting GMM in Fed-GCD setup.} We assume that the $n$-th client generates a GMM $\mathcal{G}^{L}_{n} = \{\mathcal{N}(\boldsymbol{\mu}_{i},\boldsymbol{\sigma}_{i})\}_{i=1}^{M^{L}_{n}} $ with $M^{L}_{n}$ components, where the $\boldsymbol{\mu}_{i}$ and $\boldsymbol{\sigma}_{i}$ are the mean and variance of the $i$-th component. We use a component to model a potential class/category. For simplicity, we assume that the covariance matrix $\boldsymbol{\sigma}$ is diagonal and each cluster has the equal prior probability. By maximizing the posterior of $\boldsymbol{v}_{i}$ belonging to the $y_{i}$-th cluster, the GMM loss on the $n$-th client is derived as:
\begin{equation}\label{eq:gmm}
\mathcal{L}_{gmm}(\mathcal{G}^{L}_{n}, \boldsymbol{v}_{i}, y_{i}) = -\log \frac{\left|\boldsymbol{\sigma}_{i}\right|^{-\frac{1}{2}} \mathcal{S}_{gmm}(\boldsymbol{v}_{i}, y_{i})}{\sum^{M^{L}_{n}}_{j=1, j\neq y_{i}} \left|\boldsymbol{\sigma}_j\right|^{-\frac{1}{2}} \mathcal{S}_{gmm}(\boldsymbol{v}_{j}, y_{j})},
\end{equation}

\begin{equation}\label{eq:gmm_similarity}
\mathcal{S}_{gmm}(\boldsymbol{v}, y) = \exp{\left(- \frac{1}{2}{\left\|\frac{\boldsymbol{v}-\boldsymbol{\mu}_y}{\boldsymbol{\sigma}_{y}}\right\|^2} \cdot (1+m)\right)},
\end{equation}
where the $m$ is a non-negative margin factor to increase the inter-class dispersion.
\par\noindent
\textbf{Semi-FINCH for Local GMM Initialization.} Due to the fact that the ground-truth number of classes is often unknown in practical GCD applications, we first propose to improve the parameter-free hierarchical clustering algorithm, FINCH~\cite{sarfraz2019efficient}, to a semi-supervised extension. Then, we use the improved semi-FINCH to assign pseudo labels for local data, and then estimate the cluster-specific mean and covariance to initialize the learnable GMM, as shown in~\cref{fig:framework} (b). Semi-FINCH can capture the potential semantic relationships among both labeled and unlabeled samples with the guidance of labeled data. Specifically, we search the first neighbor of the unlabeled sample by the cosine similarity, while enforcing the first neighbor of the labeled sample to be the corresponding hardest positive sample. 
\begin{table*}[ht]

\setlength\extrarowheight{-1pt}
\centering
\caption{The statistics of our Fed-GCD benchmark. We simulate different degrees of data heterogeneity in real-world Fed-GCD scenarios by adjusting the $\beta$ of parametric Dirichlet distribution to split the local training sets among clients.\label{tab:dataset}}
\resizebox{0.98\textwidth}{!}{

\begin{tabular}{c|c|cccccc|cccc}
\cmidrule[1pt]{1-12}\multirow{3}{*}{Dataset} & \multirow{3}{*}{$\beta$} & \multicolumn{6}{c|}{Client $N^{L}$=5} & \multicolumn{4}{c}{Server} \\ \cmidrule[0.5pt]{3-12}
 &  & \multicolumn{2}{c}{\# Labelled Classes} & \multicolumn{2}{c|}{\# Unlabelled Classes} & \multicolumn{2}{c|}{\# Classes Shared Across} & \multicolumn{2}{c}{Labelled} & \multicolumn{2}{c}{Unlabelled} \\ \cmidrule[0.5pt]{3-8} \cmidrule[0.5pt]{9-12}
 &  & Max & Min & Max & \multicolumn{1}{c|}{Min} &  \multicolumn{1}{c|}{all clients} & $\geq$2 clients & \#~Classes & \#~Images & \#~Classes & \#~Images \\
\cmidrule[1pt]{1-12}\multirow{2}{*}{CIFAR10~\cite{krizhevsky2009learning}} & 0.2 & 5 & 4 & 10 & 9 & \multicolumn{1}{|c|}{2} & 5& 5&2500&10&7500 \\
&0.05& 5 & 1 & 10 & 4 & \multicolumn{1}{|c|}{0} & 4&   5&2500&10&7500 \\ \hline
\multirow{2}{*}{CIFAR100~\cite{krizhevsky2009learning}} & 0.2 & 50 & 33 & 73 &100  & \multicolumn{1}{|c|}{16}&49 & 50&2500&100&7500 \\
 & 0.05 & 44 & 17 &  40&90&  \multicolumn{1}{|c|}{1}&43 &  50&2500&100&7500 \\ \hline
\multirow{2}{*}{ImageNet-100~\cite{deng2009imagenet}} & 0.2 & 50 & 37& 100&70& \multicolumn{1}{|c|}{16}&50 & 50 & 1250 & 100 & 3750  \\
 & 0.05 & 44 & 17 & 90 & 40 & \multicolumn{1}{|c|}{1}&47 &50 & 1250 & 100 & 3750\\ \hline
\multirow{2}{*}{CUB-200~\cite{wah2011caltech}} & 0.2 & 100 & 36 & 98 & 200 & \multicolumn{1}{|c|}{5}& 97& 100 & 1430 & 200 & 4362 \\
&0.05& 89 & 25 & 178 & 59 & \multicolumn{1}{|c|}{0} & 84 &   100 & 1430 & 200 & 4362 \\ \hline
\multirow{2}{*}{SCars~\cite{krause20133d}} & 0.2 & 98 & 47 & 196 & 95 & \multicolumn{1}{|c|}{5} &96 & 98 & 2001 & 196 & 6040 \\
 & 0.05 & 87 & 29 & 177 & 57 & \multicolumn{1}{|c|}{0}& 87& 98 & 2001 & 196 & 6040 \\ \hline
\multirow{2}{*}{Pet~\cite{parkhi2012cats}} & 0.2 &19  & 12 & 22 & 37&\multicolumn{1}{|c|}{3}&  19& 19&  940& 37 &2729 \\
 & 0.05 &18  &5  &43  & 16&\multicolumn{1}{|c|}{0} & 17&  19&  940& 37 &2729 \\ \cmidrule[1pt]{1-12}
\end{tabular}
}
\vspace{-1em}
\end{table*}

\par\noindent
\textbf{Connections between GCL and PCL~\cite{DBLP:conf/iclr/0001ZXH21}.}
Prototypical contrastive learning (PCL)~\cite{DBLP:conf/iclr/0001ZXH21} is a pioneering method to introduce class-level supervision into unsupervised contrastive learning. PCL estimates a scalar concentration as the temperature parameter to scale the similarity between a feature and its prototype. Although it is efficient for learning discriminative representation, it fails to model a precise representation distribution that is supposed to generate reliable representations for the downstream clustering. Here, we discuss the differences between the PCL and the GCL. The similarly metric of the PCL is given by:
\begin{equation}\label{eq:pcl_similarity}
\mathcal{S}_{pcl}(\boldsymbol{v}_{i}, y_{i}) = \exp \left(\boldsymbol{v}_{i} \cdot \boldsymbol{\mu}_{y_{i}} / \phi_{y_{i}}\right),
\end{equation}
where $\phi_i$ is the estimated temperature parameter for the $i$-th cluster. Comparing~\cref{eq:gmm_similarity,eq:pcl_similarity}, different from PCL, we model the clusters via the GMM with additional covariance matrices, and naturally derive the squared Mahalanobis distance as distance metric for contrastive learning. \textit{This allows models to dynamically control the contrastive temperatures in a dimension-wise way and to learn more reliable distributions of representations for the subsequent sampling.}
\par
Furthermore, we introduce a regularization term to explicitly compact clusters and constrain covariance, to avoid trivial solutions. For example, GMM generates a high classification accuracy, but the sample embedding is far away from the center of the cluster due to the large class-specific variance. Using the regularization loss can constrain the distance between the sample embedding and its corresponding cluster center as well as reduce the overlarge variances. The regularization loss is:
\begin{equation}\label{eq:reg}
\mathcal{L}_{reg}(\mathcal{G}^{L}_{n}, \boldsymbol{v}_{i}, y_{i}) = - \log (\mathcal{S}_{gcl}(\boldsymbol{v}_{i}, y_{i}))  + \frac{1}{2} \log \left|\boldsymbol{\sigma}_{y_{i}}\right|.
\end{equation}
Taking \cref{eq:gmm,eq:gmm_similarity,eq:reg}, the overall GCL loss is defined by a weighted sum: 
\begin{equation}\label{eq:gcl}
\mathcal{L}^{n}_{gcl}(\mathcal{G}^{L}_{n}) = \sum_{i=1}^{N_{n}^{L}}\mathcal{L}_{gmm}(\mathcal{G}^{L}_{n}, \boldsymbol{v}_{i}, y_{i})   + \alpha\mathcal{L}_{reg}(\mathcal{G}^{L}_{n}, \boldsymbol{v}_{i}, y_{i}),
\end{equation}
where $\alpha$ is a non-negative weighting coefficient. By optimizing this objective, the cluster-specific mean and variance can be learned.

\subsection{Client Semantic Association}\label{sec:csa}
In Fed-GCD task, the data distributed to clients is highly heterogeneous. Moreover, due to privacy constraints, the central server is unreasonable to get prior knowledge to align the local clusters in practical scenarios. To overcome this limitation, we propose a sample yet efficient approach, namely client semantics association (CSA). The goal of CSA is to mine common semantic knowledge from the uploaded local GMMs, and aggregate diverse local knowledge for enriching category knowledge.

\par\noindent
\textbf{Client-Agnostic Potential Semantic Association}.
Given a set of the uploaded GMMs $\mathcal{G}^{L} = \{\mathcal{G}^{L}_{n}\}^{N^{L}}_{n=1}$, we sample $N^{S}$ instances from each Gaussian distribution, which results in a new representation set. By applying unsupervised FINCH clustering on the set, the central server generates a new global GMM, as illustrated in~\cref{fig:framework}~(c). The global GMM will be sent to each client for the subsequent local training.
Intuitively, the clusters with similar semantics will be grouped into new clusters. This type of clusters can be regarded as a super-class that contains more information with a large variance. This process implicitly associates common classes scattered in clients, thereby further enriching intra-class information. On the other hand, the clusters with relatively independent semantics will be preserved. By sampling, CSA augments the category knowledge contained in global GMM, which is beneficial for providing more negative classes in the GCL on local clients. In short, by incorporating diverse knowledge from different clients, the global GMM establishes a bridge among different clients. This allows the isolated local knowledge to mutually transfer among clients, providing a complementary supervision for local GCL.

\subsection{Federated Global-Local GCL}
As illustrated in~\cref{fig:framework} (d), taking the $n$-th client as an example, we consider both the distributed global GMM $\mathcal{G}^{G}$ and the local GMM $\mathcal{G}^{L}_{n}$, to guide the optimization of the local model. We use a convex combination of them to achieve an optimal balance between the local and the global knowledge learning. The objective of GCL on the $n$-th client is:
\begin{equation}\label{eq:overall}
   \mathcal{L}^{n}=\mathcal{L}^{n}_{ins} + (1-\gamma)\mathcal{L}^{n}_{gcl} (\mathcal{G}^{G}) + \gamma \mathcal{L}^{n}_{gcl}(\mathcal{G}^{L}_{n}),
\end{equation}
where the $\gamma$ is a trade-off factor to control the strength of learning on global-local GMMs. When $\gamma$ is equal to 1, GCL leverages only the local class-level supervision for representation learning. On the contrary, GCL relies on only the aggregated global category information. 

\begin{table*}[t]
\centering
\caption{Results on generic datasets with two different degrees of data heterogeneity.\label{tab:generic}}
\resizebox{0.98\textwidth}{!}{%
\begin{tabular}{l|ccccccccc|ccccccccc}
\hline
\multirow{3}{*}{Setup} & \multicolumn{9}{c|}{NH setting~($\beta=0.2$)} & \multicolumn{9}{c}{EH setting~($\beta=0.05$)} \\ \cline{2-19} 
 & \multicolumn{3}{c}{CIFAR10} & \multicolumn{3}{c}{CIFAR100} & \multicolumn{3}{c|}{ImageNet-100} & \multicolumn{3}{c}{CIFAR10} & \multicolumn{3}{c}{CIFAR100} & \multicolumn{3}{c}{ImageNet-100} \\ \cline{2-19} 
 & All & Old & New & All & Old & New & All & Old & New & All & Old & New & All & Old & New & All & Old & New \\ \hline
\textcolor{gray}{Centralized-GCD}  & \textcolor{gray}{83.6}	&\textcolor{gray}{85.8}&\textcolor{gray}{82.0} & \textcolor{gray}{54.9}&\textcolor{gray}{56.1}& \textcolor{gray}{53.7} & \textcolor{gray}{72.1} & \textcolor{gray}{80.7 }& \textcolor{gray}{67.5} & \textcolor{gray}{83.6}	&\textcolor{gray}{85.8}&\textcolor{gray}{82.0} & \textcolor{gray}{54.9}&\textcolor{gray}{56.1}& \textcolor{gray}{53.7} & \textcolor{gray}{72.1} & \textcolor{gray}{80.7} & \textcolor{gray}{67.5} \\
\textcolor{gray}{Centralized-GCL } & \textcolor{gray}{86.7} &\textcolor{gray}{86.7} &\textcolor{gray}{86.7 }& \textcolor{gray}{58.5}&\textcolor{gray}{57.2}& \textcolor{gray}{58.1} & \textcolor{gray}{76.1} & \textcolor{gray}{83.7}& \textcolor{gray}{68.4} &\textcolor{gray}{86.7}&\textcolor{gray}{86.7 }& \textcolor{gray}{86.7} & \textcolor{gray}{58.5}&\textcolor{gray}{57.2}& \textcolor{gray}{58.1} & \textcolor{gray}{76.1} & \textcolor{gray}{83.7} & \textcolor{gray}{68.4} \\ \hline
FedAvg + GCD & 80.7 & 82.3 & 80.3 & 49.6 & 52.1 & 49.3 & 69.8 &77.1 & 65.7 & 78.7 & 80.1 & 78.3 & 47.3 & 49.2 & 45.9 & 66.4 & 74.8 & 62.1  \\
FedAvg + GCL & 83.2 & 84.9 & 82.8 & 54.1 & 55.7 & 54.0 & 74.1 & \textbf{81.8} & 67.3 & 82.2 & 82.4 & 81.9 & 52.1 & 53.2 & 51.9 & 72.5 & 79.8 & 65.3  \\ \hline
FedAvg + AGCL & 84.7 & 85.5 & 84.6 & \textbf{56.1} & \textbf{56.8} & \textbf{55.3} & \textbf{74.8} & 80.2 & \textbf{69.8} & 82.5 & 83.4 & 82.2 & 54.2 & 54.6 & 54.0 & 73.1 & 78.1 & 67.0 \\ 
FedProx + AGCL & \textbf{84.8} & \textbf{85.8} & \textbf{84.7} & 55.9 & 56.5 & 54.9 & 74.7 & 80.3 & 69.5 & \textbf{83.0} & \textbf{84.1} & \textbf{82.8} & \textbf{54.7} & \textbf{55.1} & \textbf{54.2} & \textbf{74.9} & \textbf{78.8} & \textbf{67.7}  \\ \hline
\end{tabular}
}% for resize box

\end{table*}

\section{Experiments}
\subsection{Experimental Setup}
\par\noindent\textbf{Dataset.} 
To comprehensively evaluate the performance of Fed-GCD models, we reorganize three commonly-used generic image classification datasets (\ie, CIFAR-10~\cite{krizhevsky2009learning}, CIFAR-100~\cite{krizhevsky2009learning} and ImageNet-100~\cite{vaze2022generalized}) and three more challenging fine-grained image classification datasets (\ie, CUB-200~\cite{wah2011caltech}, Stanford Cars~\cite{krause20133d}, and Oxford-IIIT Pet~\cite{parkhi2012cats}) to construct a new Fed-GCD benchmark. For each dataset, first, we sample a subset of half the classes as ``Old'' categories in the original training set, and 50\% of instances of each labeled class are drawn to form the labeled set, and all the remaining data form the unlabeled set. With the same rate of labeled-unlabeled splitting, we split the original testing set into labeled and unlabeled subsets for class number estimation and GCD testing on server. Then, we further leverage the $\beta$-Dirichlet distribution \cite{hsu2020federated} to split the training set into $N^{L}$ subsets, where the $N^{L}$ subsets are regarded as local datasets individually stored in each client. We set $N^{L}$=5 in all experiments. Experiments with different values of $N^{L}$ are studied in the supplementary materials.
\par\noindent\textbf{Evaluation Protocols.}
Due to the varying data distribution in different Fed-GCD applications, we present two evaluation protocols to separately simulate the normally heterogeneous (NH) and extremely heterogeneous (EH) scenarios by adjusting $\beta$ in Dirichlet distribution~\cite{hsu2020federated}. Specifically, we set $\beta=0.2$ and $\beta=0.05$ for NH and EH, respectively. The statistics of the dataset splits under the two evaluation protocols are described in~\cref{tab:dataset}, in which the NH setting exists few common classes but there is no labeled categories shared across all clients in the EH setting. For each dataset, we learn a global model in a decentralized training fashion. Following~\cite{vaze2022generalized}, during testing, we first estimate the number of the potential categories (\ie, $k$) in the non-overlapping test set by using the labeled data stored on server. Then we calculate the maximum of clustering accuracy between the ground truth labels and the label assignment with the estimated $k$ over the set of permutations via Hungarian algorithm~\cite{kuhn1955hungarian}. Last, we measure the clustering accuracy for ``All'', ``Old'' and ``New'' categories, respectively.

\begin{table*}[t]
\centering
\caption{Results on fine-grained datasets with two different degrees of data heterogeneity.\label{tab:fine_grained}}
\resizebox{0.98\textwidth}{!}{%
\begin{tabular}{l|ccccccccc|ccccccccc}
\hline
\multirow{3}{*}{Setup} & \multicolumn{9}{c|}{NH setting~($\beta=0.2$)} & \multicolumn{9}{c}{EH setting~($\beta=0.05$)} \\ \cline{2-19} 
 & \multicolumn{3}{c}{CUB-200} & \multicolumn{3}{c}{Stanford-Cars} & \multicolumn{3}{c|}{Oxford-Pet} & \multicolumn{3}{c}{CUB-200} & \multicolumn{3}{c}{Stanford-Cars} & \multicolumn{3}{c}{Oxford-Pet} \\ \cline{2-19} 
 & All & Old & New & All & Old & New & All & Old & New & All & Old & New & All & Old & New & All & Old & New \\ \hline
\textcolor{gray}{Centralized-GCD } & \textcolor{gray}{51.3} & \textcolor{gray}{57.3} & \textcolor{gray}{45.4} &\textcolor{gray}{39.7}& \textcolor{gray}{58.0}& \textcolor{gray}{31.2}& \textcolor{gray}{80.2} &\textcolor{gray}{85.1}& \textcolor{gray}{77.6} & \textcolor{gray}{51.3} & \textcolor{gray}{57.3} & \textcolor{gray}{45.4} &\textcolor{gray}{39.7}& \textcolor{gray}{58.0}& \textcolor{gray}{31.2}& \textcolor{gray}{80.2} &\textcolor{gray}{85.1}& \textcolor{gray}{77.6 }\\
\textcolor{gray}{Centralized-GCL}  & \textcolor{gray}{58.1}       &  \textcolor{gray}{ 55.9}     & \textcolor{gray}{ 60.3}   & \textcolor{gray}{ 41.7}   & \textcolor{gray}{55.5} &   \textcolor{gray}{38.1 } &    \textcolor{gray}{ 85.5}& \textcolor{gray}{ 85.8}  &\textcolor{gray}{  85.2} & \textcolor{gray}{58.1}         &   \textcolor{gray}{ 55.9}     &  \textcolor{gray}{60.3  }    &   \textcolor{gray}{  41.7}& \textcolor{gray}{55.5} &  \textcolor{gray}{38.1 }       &  \textcolor{gray}{   85.5}  & \textcolor{gray}{ 85.8}      & \textcolor{gray}{ 85.2}\\ \hline
FedAvg + GCD & 46.3 & 54.8 & 40.1 & 32.4 & 49.8 & 28.3&76.2& 77.8 & 75.2 & 43.3 & 52.8 & 38.9 & 30.4 & 46.1 & 26.5 & 72.1& 76.4 & 71.5  \\
FedAvg + GCL & 53.7 & 54.6 & 53.2 & 36.0 & 48.1 & 33.7 & 80.7 & 81.3 & 80.2 & 52.2 & 53.1 & 52.9 & 35.3 & 45.7 & 31.5 & 79.5 & 81.5 & 78.6 \\ \hline
FedAvg + AGCL & 55.2 & 52.5 & 56.7 & 38.2 & \textbf{50.8} & 36.0 & \textbf{82.7} & \textbf{83.9} & \textbf{82.3} & 53.1 & 52.9 & 54.2 & 36.4 & 44.9 & 32.8 & 81.4 & 82.0 & 80.7 \\ 
FedProx + AGCL & \textbf{55.4} & \textbf{52.7} & \textbf{56.8} & \textbf{38.5} & 50.7 & \textbf{36.4} & 82.5 & 83.6 & 82.2 & \textbf{53.6} & \textbf{53.2} & \textbf{54.5 }& \textbf{36.9} & \textbf{45.2} & \textbf{33.0} & \textbf{81.5} & \textbf{82.1} & \textbf{80.8} \\ \hline
\end{tabular}
}% for resize box
\vspace{-1.5em}
\end{table*}

\subsection{Implementation Details}
\par\noindent
On each client, we adopt the same backbone network, a ViT~\cite{DBLP:conf/iclr/DosovitskiyB0WZ21} pre-trained by DINO~\cite{caron2021emerging}, and use its [CLS] token for GCL learning and new category discovery. Following GCD~\cite{vaze2022generalized}, the instance contrastive learning is implemented by a projection head with 65,536 dimensions and two randomly-augmented views of an image. For a fair comparison, we follow \cite{vaze2022generalized} and set $\lambda$, $\tau^{S}$ and $ \tau^{L}$ to 0.35, 0.07 and 0.05, respectively. We fine-tune only the last block of the ViT~\cite{DBLP:conf/iclr/DosovitskiyB0WZ21} with an initial learning rate of 0.1 and upload it to the central server in each communication. The projection head and global-local GMMs are trained with an initial learning rate of 0.01. All models are optimized by SGD~\cite{qian1999momentum} for 200 epochs with a cosine annealing schedule. The size of the mini-batch is set to 128. The hyper-parameters $\alpha$, $\gamma$, $m$ and $N^{S}$ are set to 0.01, 0.9, 0.3 and 1 in all experiments.

\subsection{Performance Evaluation}
Since this work is the first to explore GCD tasks under a federated learning challenge, there is no Fed-GCD-specific method used for comparison. Moreover, to the best of our knowledge, GCD~\cite{vaze2022generalized} is the only state-of-the-art GCD method with official codes. Thus, we first adapt the GCD method into our Fed-GCD task as the strong baseline (``FedAvg + GCD''). Then, we separately implement the AGCL without global GCL (``FedAvg + GCL'') and the full AGCL (``FedAvg + AGCL'') to investigate the effects of our global-local GCL. Next, to provide a reference performance, we evaluate the centralized training performance of GCD (``Centralized-GCD'') and GCL (``Centralized-GCL''). Finally, we adapt AGCL in the advanced heterogeneous federated learning framework~\cite{DBLP:conf/mlsys/LiSZSTS20} (``FedProx + AGCL''), for a comprehensive comparison. We summarize the experimental results in~\cref{tab:generic,tab:fine_grained} and the main conclusions below.
\par\noindent
\textbf{Comparison on Generic Datasets.}
Analyzing the results in~\cref{tab:generic}, we draw two-fold conclusions: 1) A significant accuracy drop between the ``Centralized-GCD'' and the ``FedAvg + GCD'' setup, especially by 7.6\% in the EH setting on CIFAR100; 2) our AGCL consistently outperforms other setups. In the NH setting, AGCL outperforms the ``FedAvg+GCD'' by 6.5\% on CIFAR-100 for ``All'' classes. Although there is no category shared across all clients in the EH setting, AGCL still achieves consistent improvements compared with the ``FedAvg+GCD'' by 6.9\% on CIFAR-100, and 5.7\% on ImageNet-100 for ``All'' classes. 

\par\noindent
\textbf{Comparison on Fine-Grained Datasets.} The experimental results in~\cref{tab:fine_grained} show that AGCL outperforms other methods for ``All'' classes. Specifically, AGCL outperforms the baseline method on CUB-200 for ``All'' classes by 8.9\% on the NH setting and by 9.8\% on the EH setting. Compared with ``FedAvg + GCD'', AGCL shows less performance decrease under decentralized training challenge, owing to the advantage of leveraging the aggregated category information among different fine-grained clients.
\par\noindent
\textbf{Summary.} The experimental results demonstrate that 1) the proposed Fed-GCD task is challenging due to the severe data heterogeneity, which results in a large accuracy degradation between the centralized and decentralized training; 2) the fine-graded Fed-GCD exists a larger performance degradation caused by decentralized training, compared to the generic Fed-GCD. This is because the differences between different classes in fine-grained datasets are subtle and understanding the fine-grained visual is more challenging for GCD; 3) AGCL achieves consistent improvement in all settings. Benefiting from aggregating different categories scattered on clients, AGCL achieves better performance, especially on fine-grained tasks in the EH setting; 4) we verify the superiority of FedProx~\cite{li2020federated} on more heterogeneous federated learning.
\begin{table}[t]
\centering
\caption{The effectiveness of loss functions of the FedAvg based-AGCL on the EH setting ($\beta$=0.05).\label{tab:ablation}}
\resizebox{0.8\textwidth}{!}{
\begin{tabular}{c|cccc|cccccc}
\cmidrule[1pt]{1-11}
\multirow{2}{*}{Index}&\multicolumn{4}{c|}{Component}                                   & \multicolumn{3}{c}{CUB-200~\cite{wah2011caltech}} & \multicolumn{3}{c}{Oxford-Pet~\cite{parkhi2012cats}} \\ \cmidrule[0.5pt]{2-11}
 &$\mathcal{L}^{I}$ & $\mathcal{L}_{gmm}^{L}$ & $\mathcal{L}_{reg}$ &$\mathcal{L}_{gmm}^{G}$  & All          & Old          & New         & All         & Old        & New        \\ \cmidrule[1pt]{1-11}
a)&\cmark & & & & 43.3 &52.8& 38.9& 72.1& 76.4 &71.5 \\
b)& & \cmark & & & 48.9 & 50.5 & 48.5 & 76.8 & 78.5 & 75.1\\
c)& & \cmark & \cmark & & 50.6 &51.8& 49.8 & 78.0& 80.7& 77.4\\
d)&\cmark &\cmark & \cmark & & 52.2 &53.1& 52.0 & 79.5& 81.5&78.6\\
e)&\cmark & \cmark & \cmark & \cmark & 53.1 &52.9& 54.2 &81.4 &82.0& 80.7 \\ \cmidrule[1pt]{1-11}
\end{tabular}
}

\end{table}

\subsection{Effectiveness of Each Component of AGCL}\label{sec:ab}
To verify the effectiveness of each component of AGCL, we conduct five group experiments on both CUB-200~\cite{wah2011caltech} and Pet~\cite{parkhi2012cats} datasets, as shown in~\cref{tab:ablation}. The method (a) is the baseline method, \textit{i.e.}, ``FedAvg + GCD''.
\par\noindent
\textbf{Effectiveness of local GCL.}
The results of the experiment (a) and (c) indicate that GCL outperforms the baseline with instance-level supervision by a large margin, demonstrating the importance of class-level supervision in GCD. Especially for the accuracy of new classes, GCL outperforms the baseline by 10.9\% on the CUB-200 dataset. 
\par\noindent
\textbf{Effectiveness of regularization loss in~\cref{eq:reg}.} Comparing the experiment (b) and (c), we can find that enforcing the regularizing loss can achieve consistent improvement. This is because the regularization loss can encourage models to avoid trivial sub-optimal solutions.

\begin{figure}[t]
  \begin{center}
  \includegraphics[width=0.48\textwidth]{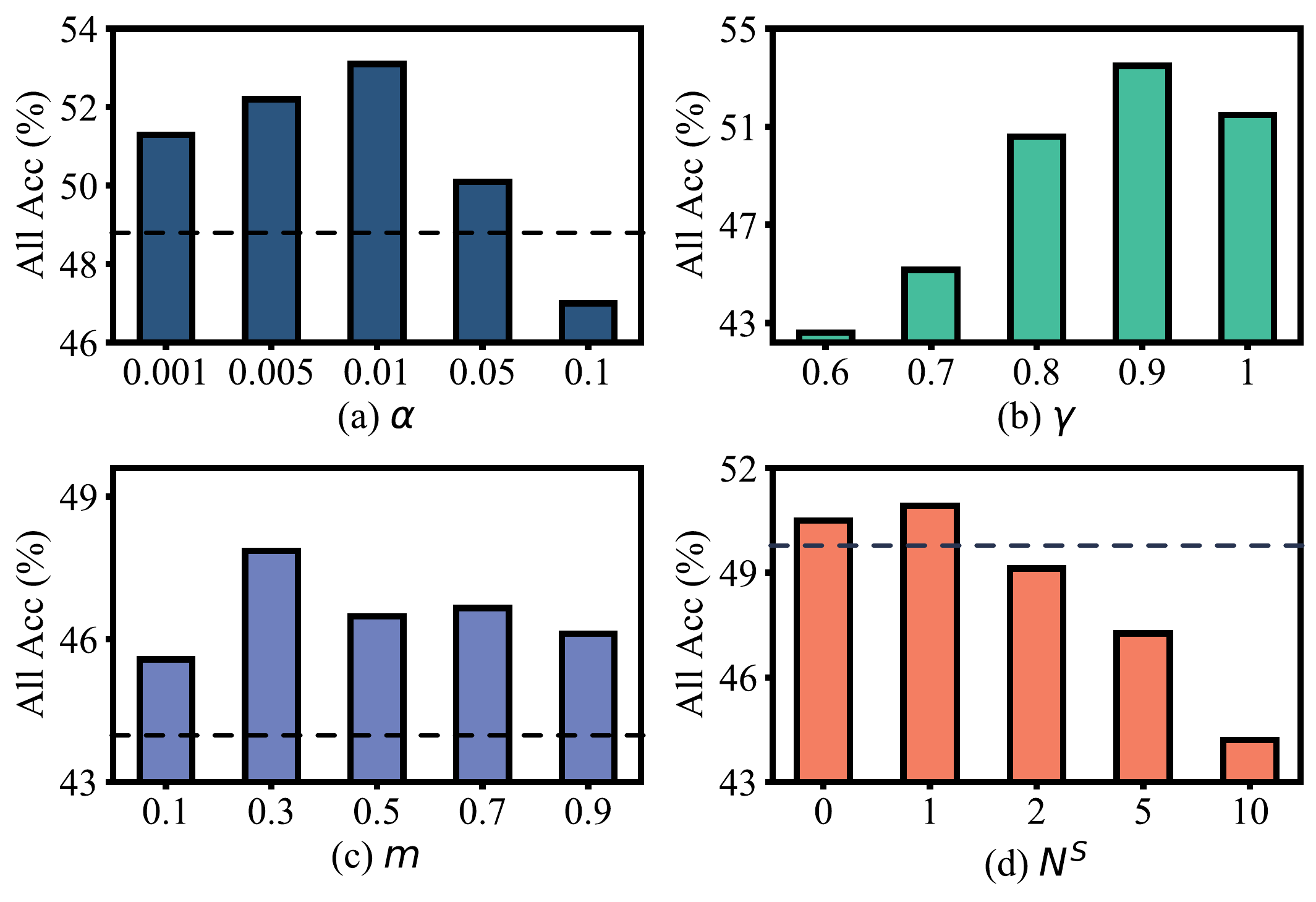}
\caption{Impact of hyper-parameters. The clustering accuracy on ``All'' categories is reported.\label{fig:hyper_parameter}}
  \end{center}

\end{figure}

\par\noindent
\textbf{Effectiveness of CSA in~\cref{sec:csa}.}
Based on the results of the experiment (d-e), we experimentally demonstrate that CSA  can associate heterogeneous category knowledge even without commonly-shared categories in EH setting. The associated knowledge contained in the global GMM complements representation learning based on local GMM, thereby improving the model's ability to discover new categories.

\subsection{Hyper-Parameter Analyses}\label{sec:hyper_parameter}
In this section, we discuss the impacts of the hyper-parameters of AGCL on the CUB-200 dataset under the EH setting, including loss weights ($\alpha$ and $\gamma$), the margin parameter of GCL loss ($m$), and the number of sampling from each potential category of local GMMs ($N^{S}$) in CSA. For each experiment, we vary the value of the studied parameter while fixing the others with default values.
\par\noindent
\textbf{Impact of regularization weight in \cref{eq:gcl}} is illustrated in~\cref{fig:hyper_parameter}~(a). A large weight may lead to worse performance compared to the configuration without the regularization loss (see the dashed line on~\cref{fig:hyper_parameter}~(a)). We empirically set $\alpha$ as 0.01 to achieve optimal performance.
\par\noindent
\textbf{Impact of trade-off factor in~\cref{eq:overall}} is illustrated in~\cref{fig:hyper_parameter}~(b). 
 
Based on the results, we find that local GCL plays a dominant role in training a discriminative representation. Meanwhile, introducing relatively few knowledge from the global GMM can complement the contrastive learning based on local GMMs. On the contrary, when AGCL mainly relies on the global GMM, the performance of local training will be largely degraded (\textit{e.g.}, $\gamma$=0.6).

\par\noindent
\textbf{Impact of margin parameter in \cref{eq:gmm_similarity}} is illustrated in~\cref{fig:hyper_parameter}~(c). We find that using margin consistently improves accuracy compared with the configuration without margin (see the dashed line on~\cref{fig:hyper_parameter}~(c)). We choose the optimal $m$=0.3 in all experiments.

\par\noindent
\textbf{Impact of sampling parameter in CSA} is illustrated in~\cref{fig:hyper_parameter}~(d). The results are summarized as: 1) our CSA can effectively aggregate heterogeneous knowledge even without sampling (see the dashed line on~\cref{fig:hyper_parameter}~(d)); 2) sampling only one sample for each category can further improve the performance; 3) sampling more samples leads to worse performance. Thus, $N^{S}$ is set to 1 in all experiments.

\section{Conclusion}
In this work, we propose a new Federated Generalized Category Discovery (Fed-GCD) task, based on the practical requirement of decentralized training trends. To handle this task, we propose a novel Associated Gaussian Contrastive Learning (AGCL) framework specifically designed to overcome the unique challenges posed by Fed-GCD. Moreover, we build a benchmark based on six visual datasets to facilitate the study of Fed-GCD. Extensive experiments show that AGCL outperforms the FedAvg-based baseline on all datasets. In future, we attempt to relieve the requirement of storing labeled data in the central server to meet more realistic  scenarios for Fed-GCD.

\newpage

\bibliography{egbib}
\bibliographystyle{iclr2022_conference}

\end{document}